%% file: bare_jrnl_new_sample4.tex
\documentclass[journal]{IEEEtran}
\usepackage{amsmath,amsfonts}
\usepackage[caption=false,font=normalsize,labelfont=sf,textfont=sf]{subfig}
\usepackage{stfloats}
\usepackage{graphicx}
\usepackage{cite}
\usepackage{multirow}
\usepackage{enumitem}
\usepackage[hidelinks]{hyperref}
\usepackage{pifont}
\usepackage[table,xcdraw]{xcolor}
\usepackage{booktabs}
\usepackage{tikz}
\usepackage{inconsolata}

\definecolor{my_green}{RGB}{51,102,0}
\definecolor{my_red}{RGB}{204, 0, 0}
\newcommand{\cmark}{\textcolor{my_green}{\ding{51}}} 
\newcommand{\xmark}{\textcolor{my_red}{\ding{55}}} 

\begin{document}

\title{Explainable Multimodal Depression Recognition in Clinical Interviews via PHQ-Aligned Symptom Summarization}

\author{Wenjie~Zheng,~\IEEEmembership{}
    Qiming~Xie,~\IEEEmembership{}
    Jianfei~Yu\IEEEauthorrefmark{1},~\IEEEmembership{}
    Yang~Wang,~\IEEEmembership{}
    Lei~Cao,~\IEEEmembership{} 
    Fei~Wang\IEEEauthorrefmark{1},~\IEEEmembership{}
    Shijin~Wang,~\IEEEmembership{}
    Rui~Xia\IEEEauthorrefmark{1},\IEEEmembership{}
    and Chengqing~Zong,~\IEEEmembership{Fellow, IEEE}

\IEEEcompsocitemizethanks{
    \IEEEcompsocthanksitem Wenjie Zheng, Qiming Xie, and Jianfei Yu are with the School of Artificial Intelligence, Nanjing University of Science \& Technology, Nanjing, China. Yang Wang and Fei Wang are with the Department of Psychiatry, the Affiliated Brain Hospital of Nanjing Medical University, Nanjing, China. Lei Cao is with the Faculty of Psychology, Beijing Normal University, Beijing, China. Shijin Wang is with the iFLYTEK AI Research (Central China), Hefei, China. Rui Xia is with the School of Intelligence Science and Technology, Nanjing University, China. Chengqing Zong is with the Institute of Automation, Chinese Academy of Sciences, Beijing, China.
    \IEEEcompsocthanksitem \IEEEauthorrefmark{1}Corresponding authors: Jianfei Yu (e-mail: jfyu@njust.edu.cn), Fei Wang (e-mail: fei.wang@yale.edu), and Rui Xia (e-mail: rxia@nju.edu.cn).
    \protect
}
}

\markboth{}{}


\maketitle

\begin{abstract}
Recent advances in multimodal depression recognition for clinical interviews (MDRC) have demonstrated the potential of AI systems by integrating textual, acoustic, and facial cues.
However, existing methods pay limited attention to interpretability, thereby constraining reproducibility and clinician review. To address this, we introduce Explain-MDRC, an explainable MDRC framework that mirrors clinical workflows by generating structured symptom summaries from text and integrating them with nonverbal cues for recognition.
Specifically, we construct Explain-DAIC, a dataset based on DAIC-WOZ and enriched with PHQ-8–aligned summary annotations, providing a foundation for developing models with built-in interpretability.
We further propose PhqCML, a model that combines PHQ-8–aligned symptom summarization with PHQ-aware contrastive learning and summary-informed multimodal fusion.
Automated metrics and expert evaluations show that Explain-MDRC improves recognition performance and provides more interpretable, clinician-readable intermediate evidence, suggesting a promising direction for transparent AI-assisted depression recognition research.
\end{abstract}


\input{section/introduction}

\input{section/related}
\input{section/dataset}

\input{section/method}
\input{section/experiments}
\input{section/discussion}
\input{section/conclusion}

\bibliographystyle{IEEEtran}
\bibliography{bibliography}

\vfill

\end{document}

%% file: section/introduction.tex
\section{Introduction}

Multimodal affective computing aims to automatically recognize and interpret human emotional states by integrating signals from multiple modalities, such as text, audio, and facial expressions~\cite{lian2024merbench,wang2025learning,li2025tracing}. These techniques have demonstrated significant potential in healthcare applications, enabling AI systems to assist in clinical assessment through the analysis of patients' behavioral and emotional cues~\cite{acosta2022multimodal,tu2025towards}.
Among mental health conditions, depression affects approximately 332 million people worldwide, yet a substantial proportion of cases remain unrecognized or undiagnosed (World Health Organization, 2024)\footnote{\url{https://www.who.int/news-room/fact-sheets/detail/depression}}. 
To address this gap, multimodal depression recognition for clinical interviews (MDRC) has been proposed to assist in the evaluation of depression severity, leveraging interviewer-participant interactions enriched with multimodal data, including textual, acoustic, and facial information. 
Given its potential to improve clinical decision‑making and enable earlier intervention, MDRC has garnered growing attention in recent years~\cite{hansen2023speech,liu2024digital}.

Most existing MDRC studies formulate the task as a classification problem, employing strategies such as signal aggregation~\cite{gong2017topic,ray2019multi}, temporal modeling~\cite{mao2022prediction,zhou2022hierarchical}, intra-modality representation learning~\cite{wu2023self,zhang2024llms}, and guidance from clinical instruments such as the Patient Health Questionnaire (PHQ)~\cite{kroenke2001phq,kroenke2009phq,nguyen2022improving,yang2023uncertainty}.
These studies have substantially advanced the state of the art in MDRC by improving predictive accuracy through well-designed frameworks.
\textbf{However}, current MDRC research remains primarily focused on recognition performance, often overlooking the need to generate human-interpretable evidence that aligns with clinical reasoning processes. 
This gap is particularly concerning, as recent studies emphasize that interpretability is essential for developing clinically transparent, human-centered, and clinician-reviewable AI systems~\cite{kundu2021ai,joyce2023explainable,savage2024diagnostic,rong2024human}. Moreover, enhancing interpretability may also strengthen recognition performance by encouraging models to ground predictions in clinically meaningful and explainable cues.

\begin{figure*}[!t]
\centering
\setlength{\belowcaptionskip}{-0.0cm}
\setlength{\abovecaptionskip}{0.1cm}
\includegraphics[width=1.0\textwidth]{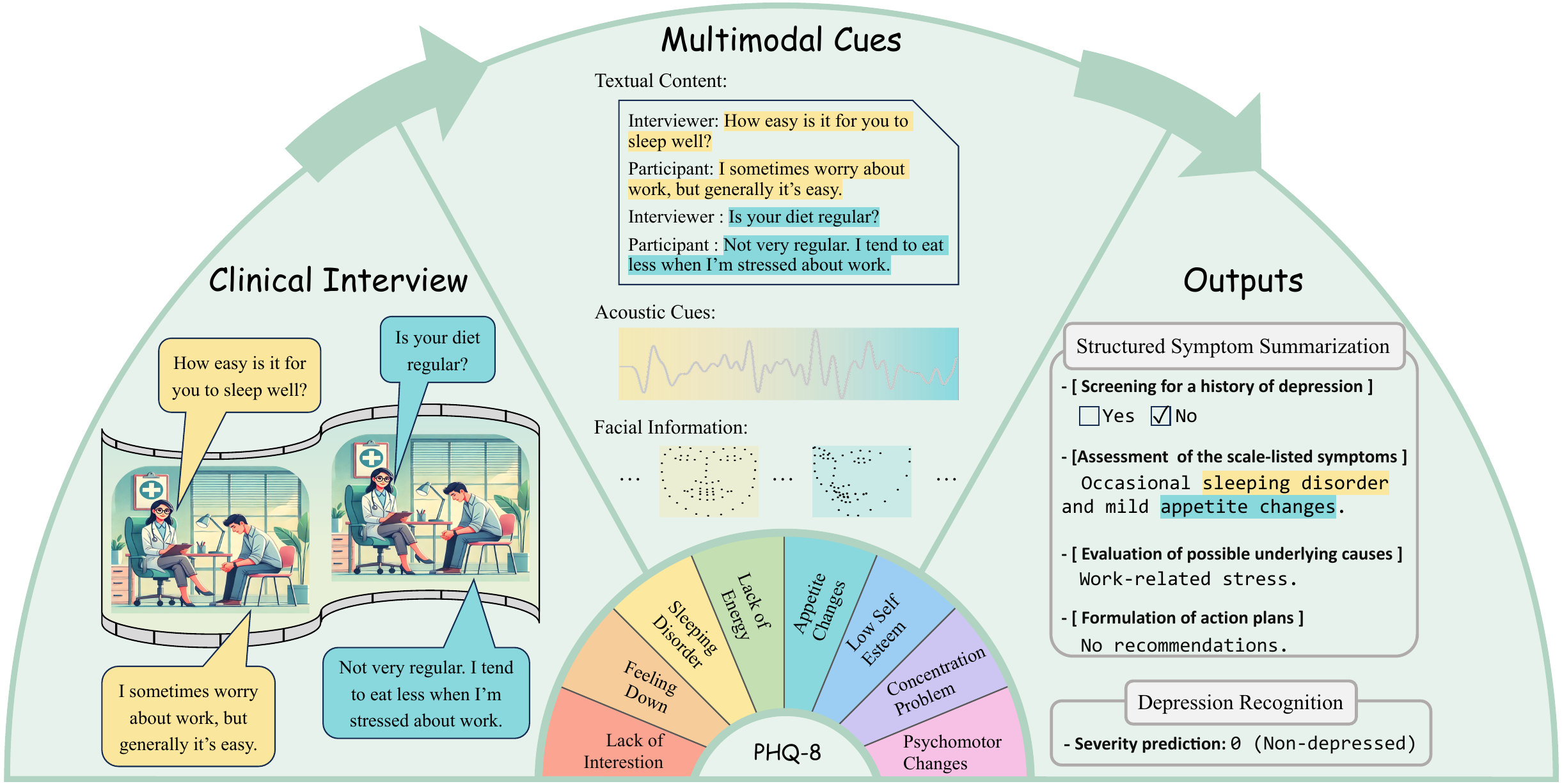}
\caption{The overview of our explainable MDRC framework, which leverages the PHQ-8 as a central pivot to provide structured and interpretable evidence for AI-assisted depression assessment. Highlighted dialogue text indicates supporting evidence for individual symptom components (e.g., yellow span = sleeping disorder).}
\label{fig:motiv}
\end{figure*}

The necessity of such interpretability becomes clear when considering the standard clinical workflow for depression. In mental health practice, clinicians typically conduct structured or semi-structured interviews to assess a patient's mental health status. 
Throughout the interview, they often evaluate symptom domains outlined in standardized depression scales and explore potential underlying causes. Importantly, clinicians synthesize this information into structured yet interpretative symptom summaries, informing clinical assessment impressions.
Moreover, nonverbal cues, such as acoustic and facial information, are observed and integrated with these verbal summaries to support a more holistic assessment.
The inability of current MDRC systems to generate such clinically meaningful summaries represents a gap between recognition capability and clinician-reviewable evidence.
Addressing this requires moving beyond classification tasks to develop models capable of producing structured, evidence-backed symptom summaries that mirror clinicians' reasoning processes. 

To this end, we introduce a framework named Explainable Multimodal Depression Recognition in Clinical Interviews (\textbf{Explain-MDRC}) that explicitly simulates core aspects of clinical workflows.
As illustrated in Fig.~\ref{fig:motiv}, given a clinical interview scenario between an interviewer and a participant, Explain-MDRC first extracts multimodal cues, including textual content, acoustic cues, and facial information. It then performs two output tasks: (1) generating a structured symptom summary aligned with PHQ-8~\cite{kroenke2009phq}, and (2) predicting the depression recognition label by integrating the generated summary with nonverbal cues.
Specifically, the structured summarization encompasses four key components aligned with standard psychiatric evaluation procedures~\cite{silverman2015apa}: \ding{182}~screening for a history of depression, \ding{183}~assessment of the scale-listed symptoms, \ding{184}~evaluation of possible underlying causes, and \ding{185}~formulation of action plans, with supporting dialogue evidence anchoring the first three components.

To operationalize the Explain-MDRC framework, we first construct \textbf{Explain-DAIC} by extending the widely used MDRC dataset DAIC-WOZ~\cite{gratch2014distress, valstar2016avec} with manually annotated PHQ-8--aligned symptom summaries.
Unlike conventional depression-recognition datasets that mainly provide interview recordings and session-level labels, Explain-DAIC explicitly links participant utterances to clinically meaningful symptom dimensions, while retaining the original textual, acoustic, and visual signals for multimodal recognition.
This design supports both clinician-readable symptom explanation generation and the evaluation of whether such explanations improve downstream depression recognition.

Building on this dataset, we propose \textbf{PhqCML} (PHQ-aware Contrastive Multimodal Learning), a trainable two-stage method that makes the symptom summary an explicit intermediate reasoning layer rather than a post-hoc explanation.
In the first stage, PhqCML instruction-tunes an LLM to generate structured symptom summaries, and uses PHQ-aware contrastive learning to align each summary's hidden representation with the participant's eight-dimensional PHQ-8 profile during training.
In the second stage, PhqCML fuses the resulting PHQ-aware summary representation with acoustic and visual features for depression recognition.

Experiments on the Explain-DAIC dataset support the effectiveness of the proposed method.
Specifically, PhqCML outperforms the previous state-of-the-art by 7.10 absolute percentage points in Macro-F1, underscoring that incorporating structured symptom summaries not only enhances recognition performance but also improves model interpretability.
In a clinical expert evaluation, PhqCML outperformed GPT-5.5 on completeness, correctness, and conciseness, with all three comparisons reaching $p<0.001$ under case-level one-sample Wilcoxon signed-rank tests, indicating improved clinical readability of the generated symptom summaries.

The main contributions are summarized as follows:
\ding{182} We formulate Explain-MDRC, an explainable MDRC framework that couples PHQ-8-aligned symptom summarization with multimodal depression recognition, making clinician-readable symptom evidence an explicit intermediate output.
\ding{183} We construct Explain-DAIC, a DAIC-WOZ-derived dataset enriched with manually annotated PHQ-8-aligned symptom summaries, enabling joint study of symptom explanation generation and interview-based depression recognition.
\ding{184} We propose PhqCML, which combines PHQ-aware contrastive learning and summary-informed multimodal fusion, achieving stronger recognition performance while providing structured intermediate evidence in clinical interviews.

Collectively, this work moves MDRC from black-box classification toward clinician-aligned symptom reasoning for AI-assisted depression recognition research.

\input{table/data_statis}

%% file: table/data_statis.tex
\begin{table*}[!t]
\centering
\small
\setlength{\belowcaptionskip}{0.1cm}
\setlength{\abovecaptionskip}{0.2cm}
\caption{\label{tab:data_statis}Comparison of the proposed Explain-DAIC dataset with mainstream multimodal depression recognition datasets for clinical interviews. The turns in parentheses denote the total number of question--response units in the dataset. ``Sym. Sum.'' refers to the symptom summarization task and ``Dep. Rec.'' corresponds to the depression recognition task.
}
\setlength{\tabcolsep}{1.6mm}
\begin{tabular}{lcrrrccc}
\toprule
Dataset  & Modality & \# Subjects & \# Dialogues (Turns)  & Duration & Language  & Sym. Sum. & Dep. Rec.  \\
\midrule
CMDC (2022)~\cite{zou2022semi}      & text, audio, video      & 78      & 78 (1,848)  &    46.0h & Mandarin   & \xmark   &\cmark      \\
EATD-Corpus (2022)~\cite{shen2022automatic}   & text, audio   & 162     & 162 (486)   & 2.3h & Mandarin  & \xmark   &\cmark       \\
DAIC-WOZ (2016)~\cite{valstar2016avec}    & text, audio, video      & 189     & 189 (21,484)   & 47.3h & English   & \xmark   &\cmark  \\
\textbf{Explain-DAIC (Ours)}     &text, audio, video     &189    &189 (21,484)  & 47.3h & English   & \cmark  & \cmark \\
\bottomrule
\end{tabular}
\end{table*}

%% file: section/related.tex
\section{Related Work}



\subsection{Depression Recognition Datasets}
\label{dep_rec_dataset_survey}
Depression recognition aims to assess depressive severity from linguistic, acoustic, visual, or multimodal behavioral cues. Early text-oriented resources provide symptom-level supervision from social media or self-reported mental-health descriptions, including PHQ-9-aligned tweets~\cite{yadav2020identifying}, PRIMATE~\cite{gupta2022learning}, PsySym~\cite{zhang2022symptom}, and BDI-Sen~\cite{perez2023bdi}. These datasets annotate symptoms at tweet, sentence, or question-response levels, enabling symptom-aware text modeling and fine-grained linguistic analysis. However, because they lack synchronized acoustic and visual behavior, they are not designed for multimodal clinical interview assessment.
As research moved toward more ecologically valid scenarios, multimodal depression recognition datasets emerged across social media, in-the-wild videos, controlled tasks, and clinical interviews. Social-media datasets include multiTwitterDep~\cite{gui2019cooperative}, multiRedditDep~\cite{uban2022explainability}, and the meme-level RESTORE dataset with fine-grained symptom labels~\cite{yadav2023towards}; in-the-wild resources collect video-level samples from public platforms~\cite{yoon2022d,sawadogo2024ptsd,he2025lmvd}; and task-oriented corpora such as AVEC 2013/2014~\cite{valstar2013avec,valstar2014avec} and MODMA~\cite{cai2022multi} record participants under controlled protocols. These resources broaden the sensing context from written self-expression to speech, facial behavior, and task responses, but their labels are usually user-, video-, or session-level. Clinical-interview datasets, including CMDC~\cite{zou2022semi}, EATD-Corpus~\cite{shen2022automatic}, and DAIC-WOZ, capture richer doctor-patient or interviewer-participant interactions.

\subsection{Depression Recognition Methods}
Existing depression recognition methods improve prediction by introducing clinically informed supervision and richer behavioral representations. Early PHQ-aware studies use screening interviews, questionnaire responses, or symptom spans as intermediate signals for depression prediction. Rinaldi et al.~\cite{rinaldi2020predicting} model interview prompts with latent categories, Nguyen et al.~\cite{nguyen2022improving} leverage questionnaire-response prediction to improve generalization, and Zhang et al.~\cite{zhang2023phq} learn symptom-related spans with PHQ-aware contrastive objectives. Other work broadens the modeling scope by learning shared and disorder-specific cues across mental disorders~\cite{chen2023detection} or by capturing local topic patterns and global temporal context for severity prediction~\cite{huang2024rethinking}.

With the rapid advancement of LLMs, several studies have begun integrating LLMs into mental-health assessment pipelines. Yang et al.~\cite{yang2024mentallama} and Xu et al.~\cite{xu2024mental} build domain-specialized mental-health LLMs based on foundation models such as LLaMA and Alpaca~\cite{taori2023stanford}, while Yang et al.~\cite{yang2024psychogat} and Hu et al.~\cite{hu2026agentmental} explore interactive agent-based psychometric assessment. These studies improve task coverage and interaction ability, yet they mainly target broad mental-health prediction, screening, or adaptive questioning.
Recent interview-oriented methods further use LLMs to extract depression indicators or estimate rating-scale scores from clinical conversations~\cite{sadeghi2024harnessing,li2025spokenllm,kebe2025llamadrs}, suggesting that language models can recover clinically meaningful cues from long interviews. Multimodal systems such as IDRL~\cite{wang2026idrl}, PsyGAT~\cite{vyalla2026psychologically}, and dynamic summary generation~\cite{teng2026dynamic} also improve individual-aware, psychologically grounded, or summary-guided reasoning. Despite these advances, most MDRC systems still focus on final depression labels or task-specific explanations, rather than building PHQ-8-aligned, evidence-grounded symptom summaries as an explicit intermediate clinical representation.

\subsection{Medical Dialogue Summarization}

In the general medical field~\cite{zhang2024generalist}, Krishna et al.~\cite{krishna2021generating} employ a T5-based model to generate structured clinical summaries from physician–patient conversations. Michalopoulos et al. \cite{michalopoulos2022medicalsum} introduce MedicalSum, a framework for summarizing medical conversations by integrating medical domain knowledge from the Unified Medical Language System.
With the rise of LLMs, Van et al. \cite{van2024adapted} evaluate eight adapted LLMs across four clinical summarization tasks and demonstrated that, in many cases, these models can generate summaries comparable to or surpassing those produced by medical experts. 
Ghosh et al. \cite{ghosh2024clipsyntel} introduce the MMQS multimodal dataset and propose a CLIP- and LLM-based framework that integrates visual cues with textual queries to generate more accurate and medically informed question summaries.
Saab et al. \cite{saab2024capabilities} introduce Med-Gemini, whose strong multimodal and long-context reasoning capabilities enable it to exceed human experts in clinical text summarization.
Yuan et al. \cite{yuan2025lcds} propose LCDS, a logic-controlled system that reduces hallucinations in discharge summary generation through source-summary mapping, rule-based constraints, and expert review.
In the mental health domain, Yao et al. \cite{yao2022d4} introduce D$^4$, a Chinese depression-diagnosis–oriented dialogue dataset grounded in ICD-11 and DSM-5, supporting multiple clinical tasks and improving both empathy and diagnostic accuracy in consultation dialogue systems.
Aragón et al. \cite{aragon2024delving} develop a method that extracts BDI-related evidence from social media posts and uses an LLM to estimate answers to the 21 BDI-II items, achieving competitive performance with expert assessments.


%% file: section/dataset.tex
\section{Dataset}
\label{sec:dataSource}

\subsection{Data Source}
Since no existing MDRC dataset provides the structured symptom summaries required to develop the Explain-MDRC framework, we construct a new dataset by extending DAIC-WOZ, a widely used MDRC dataset of clinical interviews with veterans in which a clinician-controlled virtual agent (Ellie) conducts semi-structured interviews. Each participant provided informed consent and completed the PHQ-8 prior to the session.

\subsection{Motivation for Explain-DAIC Annotation}
In clinical interview settings, semi-structured interviews do not strictly follow the PHQ-8; clinicians interpret spontaneously disclosed information and integrate nonverbal cues to inform clinical assessment impressions. \textbf{However}, DAIC-WOZ lacks structured annotations of participants' specific depressive symptoms and underlying causes, which limits its use for interpretable modeling.
To address this gap, we use the PHQ-8 item structure as an annotation scaffold for organizing dialogue-grounded symptom evidence, while retaining the original DAIC-WOZ binary depression labels as the classification target.
This yields Explain-DAIC, a dataset of PHQ-8-aligned symptom summaries supporting more transparent AI-assisted depression recognition research without redefining the original DAIC-WOZ recognition task.

\begin{figure*}[htb]
\centering
\setlength{\belowcaptionskip}{0.1cm}
\setlength{\abovecaptionskip}{0.2cm}
\includegraphics[width=\textwidth]{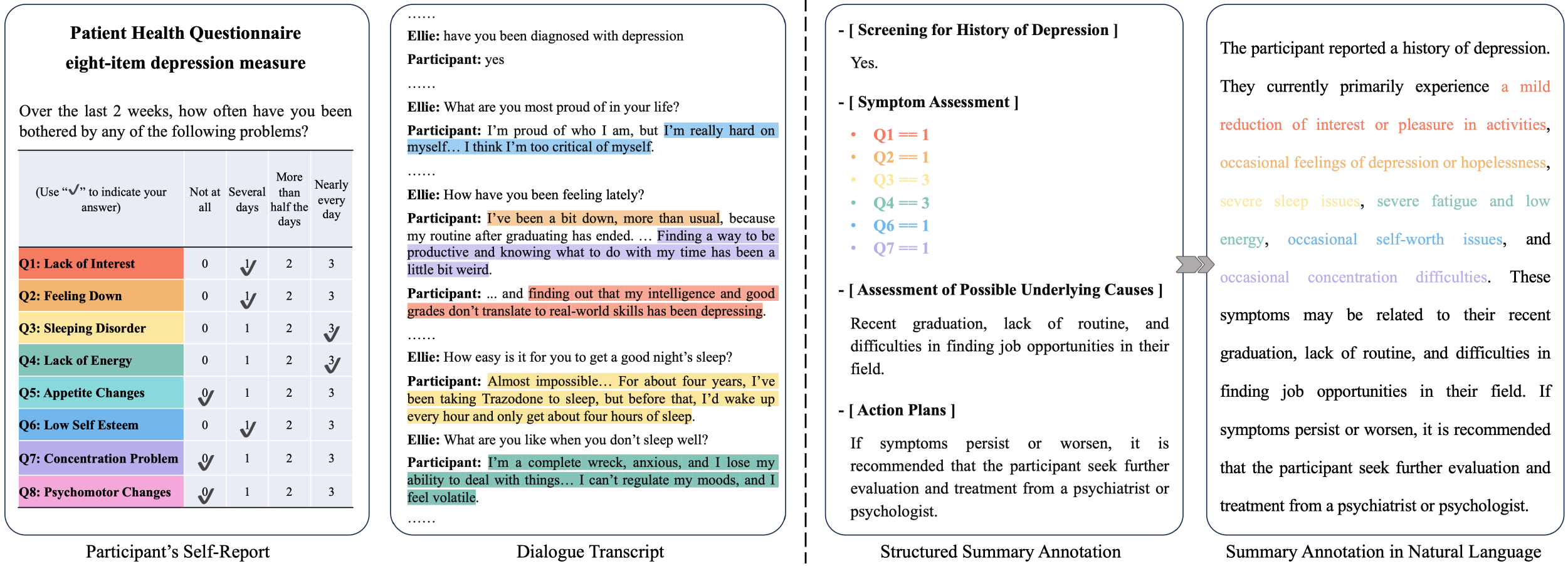}
\caption{Annotation process for a participant's structured symptom summary. For each symptom reported in the PHQ-8, annotators identify corresponding dialogue segments and assess possible underlying causes. Additional symptoms emerging during the conversation (such as Concentration Problem in this example) are also included in the summary.}
\label{fig:annot}
\end{figure*}

\subsection{Symptom Summarization Annotation}
\label{sec:sumAnno}
We employ three clinically trained mental health professionals who were supervised by a licensed psychiatrist to annotate structured symptom summaries. To support the development of an explainable framework, annotators are instructed to construct symptom summaries based on the interviewer--participant dialogue transcripts from the DAIC-WOZ dataset.

\textbf{Annotation Guidelines.}
Following discussions with the licensed psychiatrist, we define four required annotation components based on dialogue content.
First, annotators are required to determine whether the participant has a history of depression.
Second, annotators identify the participant's depressive symptoms, using
  the original DAIC-WOZ PHQ-8 item scores as a reference for the symptom
  dimensions to be examined, rather than as direct text to be copied into the
  summary. For each symptom indicated by the PHQ-8, annotators must locate
  corresponding explanatory evidence spans from the dialogue whenever possible;
  item-level symptom descriptions may also be adjusted when the dialogue
  clearly contradicts the self-reported score (e.g., a participant exhibits
  \textit{occasional concentration difficulties} during the interview not
  reflected in the self-report).
  Third, annotators extract supporting evidence from the dialogue to infer
  potential underlying causes of the identified symptoms.
  Finally, based on the assembled dialogue-grounded symptom evidence, annotators
  provide conservative screening-oriented recommendations. Together, these
  four components constitute the final structured symptom summary.

\textbf{Summarization Format.}
Each symptom summary consists of four components:
\ding{182}~Screening for a history of depression. If Ellie asks the participant whether they have been diagnosed with depression and the participant responds ``\textit{Yes},'' the first sentence begins with: ``\textit{The participant reported a history of depression.}'' \ding{183}~Assessment of the scale-listed symptoms. Conversely, if the participant does not have a history of depression, the summary starts directly with the second sentence: ``\textit{The participant primarily experiences [symptoms].}'' \ding{184}~Evaluation of possible underlying causes. The third sentence provides an explanation for the onset or exacerbation of these symptoms, such as: ``\textit{These symptoms may be related to their [underlying causes].}'' \ding{185}~Formulation of action plans. This component provides a conservative screening-oriented recommendation for clinician-readable reporting, for example: ``\textit{If symptoms persist or worsen, it is recommended that the participant seek further professional evaluation from a psychiatrist or psychologist.}''

\textbf{Annotation Procedure.}
As illustrated in Fig.~\ref{fig:annot} (a), each annotator receives the participant's self-reported PHQ-8 questionnaire results and the corresponding semi-structured interview transcript for constructing the reference annotations. Annotators independently perform structured annotation following the established guidelines.
The structured annotations are then converted into coherent natural language descriptions, which serve as explanatory evidence for downstream depression recognition models.

\textbf{Annotation Consistency Verification.}
For certain symptoms, such as \textit{lack of energy}, the severity to which they are expressed by the participant in dialogue can be ambiguous, and the underlying causes may be interpreted differently due to annotators' subjective judgments.
To account for this variability, three trained annotators independently annotate symptom summaries, and we adopt Fleiss' $\kappa$~\cite{fleiss1971measuring} to measure the inter-annotator agreement.
Two annotations are considered consistent if their Intersection-over-Union (IoU) score exceeds $0.5$, following the partial-match convention widely used in NLP span annotation~\cite{uzuner2011i2b2}.

We report $\kappa$ in two stages. The pre-consensus first-pass $\kappa = 0.71$ reflects raw agreement before any re-annotation, which corresponds to substantial agreement under the Landis--Koch scale~\cite{landis1977measurement}. Samples below the consistency threshold are re-annotated through structured disagreement resolution until the required level is achieved; in total, $19.1\%$ ($9/47$) of dialogues undergo this adjudication. The post-consensus $\kappa$ rises to $\mathbf{0.84}$, indicating near-perfect agreement.
We additionally report per-component $\kappa$ across the four summary fields: history of depression ($\kappa_h = 0.91$), symptom items ($\kappa_s = 0.82$), contextual stressors ($\kappa_c = 0.79$), and action plan ($\kappa_a = 0.93$). Agreement is highest on structured, factual fields (history, action plan) and lowest on contextual stressors, where subjective judgment about what constitutes a stressor introduces unavoidable variability, consistent with the ambiguity illustrated by the \textit{lack of energy} example above.

\begin{figure*}[htb]
\centering
\setlength{\belowcaptionskip}{-0.1cm}
\setlength{\abovecaptionskip}{0.1cm}
\includegraphics[width=0.98\textwidth]{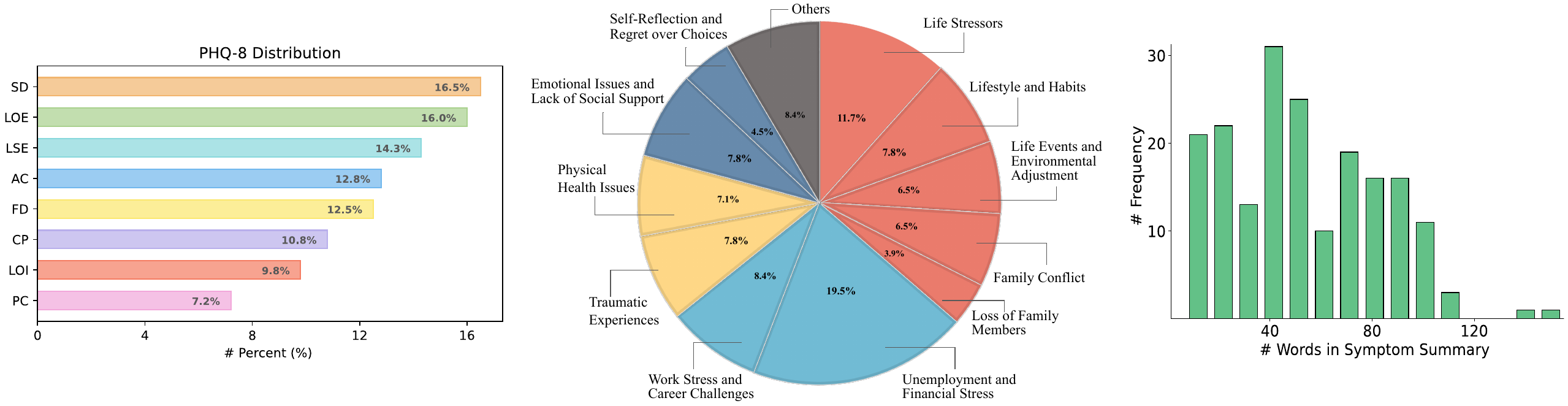}
\caption{Analysis of the annotated symptom summaries in Explain-DAIC. \textbf{Left:} PHQ-8 symptom mention distribution (\textit{SD}: Sleeping Disorder, \textit{LOE}: Lack of Energy, \textit{LSE}: Low Self-Esteem, \textit{AC}: Appetite Changes, \textit{FD}: Feeling Down, \textit{CP}: Concentration Problem, \textit{LOI}: Lack of Interest, \textit{PC}: Psychomotor Changes). \textbf{Middle:} Frequently identified causes. \textbf{Right:} Symptom summary length distribution.}
\label{fig:annotation_analysis}
\end{figure*}

\subsection{Dataset Analysis}
\label{sec:dataAnaly}
Fig.~\ref{fig:annotation_analysis} analyzes the annotated symptom summaries in Explain-DAIC.
The \textbf{left panel} shows the distribution of PHQ-8 symptom mentions across participants, with \textit{SD} (Sleeping Disorder) and \textit{LOE} (Lack of Energy) appearing most frequently.
This pattern is consistent with the clinical observation that sleep disturbance and fatigue are common symptoms in depression-related interviews, including cases with mild or subthreshold severity.
The \textbf{middle panel} summarizes the frequently identified contextual causes: Unemployment and Financial Stress (19.5\%), Life Stressors (11.7\%), and Work Stress and Career Challenges (8.4\%) are the top three categories, collectively accounting for 39.6\%.
These results suggest that economic, life-event, and occupational stressors constitute the dominant contextual factors in Explain-DAIC.
The \textbf{right panel} shows that most symptom summaries contain fewer than 80 words, reflecting the dataset's predominance of minimal-to-mild depressive symptoms while still leaving room for longer summaries when participants report more complex symptom profiles or underlying causes.

\subsection{Relationship to Original DAIC-WOZ Labels}
\label{sec:label_consistency}
Explain-DAIC augments DAIC-WOZ with PHQ-guided, dialogue-grounded symptom evidence profiles, but does not redefine the original depression-recognition task. We retain the original DAIC-WOZ binary labels for downstream recognition to preserve comparability with prior MDRC studies. The PHQ-8 item scores are used as annotation references and training-time supervision for PHQ-aware representation learning; at inference time, the PHQ-8 item vector is not used as an input feature. Thus, the task is framed as interview-based prediction of a standardized screening-derived label, rather than independent clinical diagnosis.

The refined symptom-evidence totals remain highly aligned with the original self-reported PHQ-8 totals (Pearson $r=0.985$). If these refined item-level scores were summed and thresholded at the conventional PHQ-8 cutoff, only 6 of 189 participants would cross the binary boundary. We therefore keep the original DAIC-WOZ labels unchanged and treat these cases as construct-discordant examples between self-reported screening scores and interview-derived evidence.

%% file: section/method.tex
\section{Methodology}

\subsection{Formulation of the Proposed Task}
Each clinical interview sample contains an interviewer-participant dialogue and the participant's nonverbal acoustic and visual cues. Let $u_i = (u_{i}^{t}, u_{i}^{a}, u_{i}^{v})$ denote the multimodal input of the $i$-th interview, where $u_{i}^{t}$ is the dialogue transcript, $u_{i}^{a}$ is the audio recording, and $u_{i}^{v}$ denotes the participant's visual information. Conventional MDRC aims to predict the binary depression label $y_i$ directly from these multimodal cues:
\begin{equation}
\hat{y}^{\text{DR}}_{i} = f_{\text{MDRC}}(u_{i}^{t}, u_{i}^{a}, u_{i}^{v}),
\end{equation}
where $\hat{y}^{\text{DR}}_{i} \in \{\text{depressed}, \text{non-depressed}\}$.
Following the standard DAIC-WOZ protocol, the binary label is defined by the original self-reported PHQ-8 cutoff. Accordingly, Explain-MDRC is framed as interview-based prediction of a standardized screening label, coupled with PHQ-8-aligned symptom summarization, rather than independent clinical diagnosis.

Explain-MDRC extends this formulation by introducing a clinician-readable intermediate output. Let $\mathcal{D} = \{(u_i, s_i, y_i)\}_{i=1}^{N}$ denote the Explain-DAIC corpus, where $s_i$ is the PHQ-8-aligned structured symptom summary annotated from the transcript. Given an interview, Explain-MDRC first generates a PHQ-8-aligned symptom summary:
\begin{equation}
\hat{s}_i = g_{\text{sum}}(u_{i}^{t}),
\end{equation}
and then predicts the depression label by combining the generated summary with nonverbal multimodal cues:
\begin{equation}
\hat{y}^{\text{DR}}_{i} = f_{\text{Explain-MDRC}}(\hat{s}_i, u_{i}^{a}, u_{i}^{v}),
\end{equation}
where $\hat{s}_i$ serves as explicit symptom-level evidence associated with the final prediction. Thus, the proposed task couples PHQ-8-aligned symptom summarization with multimodal depression recognition, making the explanatory evidence part of the prediction pipeline rather than a post-hoc interpretation.

\input{figure/framework_tex}

\subsection{Framework Overview} 
As shown in Fig.~\ref{fig:framework}, PhqCML adopts a unified two-stage architecture. 
The first stage trains an LLM to produce a PHQ-8-aligned symptom summary,
with PHQ-aware contrastive learning (PhqCL) using each participant's
eight-dimensional PHQ score vector as training-time clinical supervision on
the summary representation, so that summaries from participants with similar
PHQ-8 profiles are pulled together and those with dissimilar profiles pushed
apart in the LLM's hidden space.
In the second stage, the resulting PHQ-aware summary is fused with acoustic and visual features in an LLM-based architecture to produce the final depression prediction.

\subsection{Stage 1: PHQ-aware Symptom Summarization}
\label{sec:phq_aware_summary}

In clinical practice, depression assessment is anchored to a small set of standardized symptom dimensions, most commonly the eight items of the PHQ-8 scale. A useful symptom summary should read as clear clinical text and faithfully reflect a participant's profile across these eight dimensions. We designed the first stage of Explain-MDRC around this dual requirement.
 
A natural starting point is to fine-tune an LLM to generate the structured summary using the standard negative log-likelihood (NLL) objective. This objective shapes what the summary says at the token level, but it does not constrain where the resulting summary lies in the model's internal representation space. 
In practice, this means that two participants with very different PHQ-8 profiles can produce summaries whose internal representations are almost indistinguishable. 
Such collapse weakens the subsequent multimodal depression recognition stage, where the summary representation is fused with acoustic and visual cues, and it also blurs the clinical signal that the summary is intended to carry.
 
To mitigate this weakly structured representation space, we treat each summary as a point in an eight-dimensional clinical space defined by the PHQ-8 scale during training. We then add a geometric training objective: it encourages summary representations from participants with similar PHQ-8 profiles to be close, while separating representations from clinically dissimilar profiles through the contrastive denominator. The first-stage model is trained with two complementary signals: a standard NLL loss that supervises summary generation at the surface level, and a PHQ-aware contrastive loss, denoted $\mathcal{L}_{\text{PhqCL}}$, that supervises the clinical geometry of the underlying representation.

\vspace{3pt}

\subsubsection{\textbf{Symptom Summary Generation}}

We employ an LLM as the generation backbone and instruction-tune it to produce structured symptom summaries that follow the four-component clinical format introduced in Section~\ref{sec:sumAnno}. The generated summary is delimited by the markers \texttt{<summary>} and \texttt{</summary>}, which also serve as anchors for representation extraction in the contrastive objective described below.

Formally, given a participant's interview dialogue $D_i = \{d_1, \dots, d_n\}$ and an instruction prompt $I$ as input, the LLM generates the symptom summary $s_i$ autoregressively:
{\small
\begin{equation}
    s_i = \text{LLM}_{\theta}(I \oplus D_i),
\end{equation}
}
where $\oplus$ denotes text concatenation and $\theta$ denotes the trainable parameters. The generation objective is the standard token-level NLL loss:
{\small
\begin{equation}
    \mathcal{L}_{\text{SymSum}} = -\frac{1}{N}\sum_{i=1}^{N}\sum_{j=1}^{|s_i|} \log P_{\theta}\bigl(s_{i,j} \mid s_{i,<j}, \, I \oplus D_i\bigr),
\end{equation}
}
where $s_{i,j}$ denotes the $j$-th token of the target summary and $s_{i,<j}$ denotes all preceding tokens.

\vspace{3pt}
\subsubsection{\textbf{PHQ-aware Contrastive Learning}}

The NLL objective shapes the model's output distribution at the token level, but does not constrain how participants' summarization representations are organized in the LLM's hidden space. We therefore introduce a contrastive objective that operates directly on summarization representations, supervised during training by participants' eight-dimensional PHQ-8 score vectors $\mathbf{p}_i \in \{0,1,2,3\}^{8}$. 
A representation space that respects this clinical proximity provides more clinically informative features for the subsequent multimodal recognition stage.

Summarization representation.
For each training example $i$, we extract the final-layer LLM hidden states corresponding to the generated symptom-summarization span, apply mean pooling, and map the pooled vector into a contrastive embedding space using a two-layer MLP projection head $g_{\phi}$ followed by $\ell_2$ normalization:
{\small
\begin{equation}
    \mathbf{z}_i =
    \frac{
    g_{\phi}\bigl(\mathrm{MeanPool}(\mathbf{h}_i^{(\mathrm{sum})})\bigr)
    }{
    \left\lVert
    g_{\phi}\bigl(\mathrm{MeanPool}(\mathbf{h}_i^{(\mathrm{sum})})\bigr)
    \right\rVert_2
    }
    \in \mathbb{S}^{d-1},
    \label{eq:summary_repr}
\end{equation}
}
where $\mathbf{h}_i^{(\mathrm{sum})}$ denotes the final-layer hidden states over the summarization span, and $\mathbb{S}^{d-1}$ is the unit hypersphere in $\mathbb{R}^{d}$.

Contrastive learning with continuous PHQ supervision (PhqCL).
These PHQ-8 profiles provide a structured clinical coordinate system for measuring participant similarity. We therefore encourage representations to reflect graded proximity in the PHQ-8 space, so that participants with similar symptom profiles are pulled closer while clinically dissimilar profiles are separated through the contrastive denominator. For each anchor $i$ and candidate $j$, we define a clinical similarity weight
{\small
\begin{equation}
    w_{ij} = \exp\!\left(-\,\frac{\bigl\lVert \mathbf{p}_i - \mathbf{p}_j \bigr\rVert_2}{\tau_{\text{PHQ}}}\right) \;\in\; (0, 1],
    \label{eq:phq_weight}
\end{equation}
}
where $\tau_{\text{PHQ}}$ controls the decay rate of clinical similarity. This formulation replaces hard positive-pair assignment with graded PHQ-8 similarity, making PhqCL a soft clinical extension of supervised contrastive learning~\cite{khosla2020supervised}.

The PHQ-aware contrastive loss is then a soft-weighted InfoNCE:
{\small
\begin{equation}
    \mathcal{L}_{\text{PhqCL}} = -\,\frac{1}{|\mathcal{B}|}\sum_{i \in \mathcal{B}} \frac{\displaystyle\sum_{j \in \mathcal{N}(i)} w_{ij} \,\log \frac{\exp\bigl(\mathbf{z}_i \!\cdot\! \mathbf{z}_j / \tau\bigr)}{\sum_{k \in \mathcal{N}(i)} \exp\bigl(\mathbf{z}_i \!\cdot\! \mathbf{z}_k / \tau\bigr)}}{\displaystyle\sum_{j \in \mathcal{N}(i)} w_{ij}},
    \label{eq:l_phq}
\end{equation}
}
where $\mathcal{B}$ is the mini-batch, $\mathcal{N}(i)$ excludes the anchor, and $\tau$ is the contrastive temperature. Candidates with higher PHQ-8 similarity receive larger positive weights, while clinically distant candidates contribute mainly through the contrastive denominator.

\subsubsection{\textbf{Joint optimization}}
We jointly optimize symptom summarization generation and PHQ-aware representation alignment with the combined loss:
\begin{equation}
    \mathcal{L} = \mathcal{L}_{\text{SymSum}} + \lambda \cdot \mathcal{L}_{\text{PhqCL}},
    \label{eq:total_loss}
\end{equation}
where $\lambda$ is a trade-off hyperparameter.

\subsection{Stage 2: Multimodal Depression Recognition}

After obtaining the PHQ-aware symptom summaries, we perform multimodal depression recognition by fusing textual, acoustic, and visual representations within a unified LLM architecture. We first extract unimodal embeddings $\{\mathbf{X}^t, \mathbf{X}^a, \mathbf{X}^v\}$ from the three modalities:

\begin{itemize}[leftmargin=0.12in]
    \item Textual: We extract symptom-aware textual embeddings from the LLM hidden states corresponding to the generated symptom summary $s_i$. We use the final transformer layer hidden states $\mathbf{X}^{t} \in \mathbb{R}^{l_t \times d_t}$ as the textual representation, where $l_t$ denotes the summary token length and $d_t$ is the LLM hidden dimension. These embeddings are extracted offline from the first-stage LLM and projected into the second-stage LLM's embedding space via a linear projection layer.
    \item Acoustic: We adopt Wav2Vec2\footnote{\url{https://huggingface.co/jonatasgrosman/wav2vec2-large-xlsr-53-english}}, fine-tuned on Common Voice 6.1, to obtain frame-level acoustic embeddings $\mathbf{X}^{a} \in \mathbb{R}^{l_a \times d_a}$, where $l_a$ denotes the number of audio frames.
    \item Visual: Given a 3D facial landmark sequence $\mathcal{F} = (f_1, \dots, f_n)$, we employ PointMLP~\cite{marethinking} to extract visual embeddings $\mathbf{X}^{v} \in \mathbb{R}^{l_v \times d_v}$, where $l_v$ denotes the length of the facial sequence.
\end{itemize}

Since the acoustic and visual sequences are much longer than the textual summary, directly feeding them into the LLM would be computationally prohibitive. We therefore employ a Perceiver Resampler-based modality encoder~\cite{alayrac2022flamingo} for each non-textual modality, which compresses variable-length feature sequences into a fixed set of $N_m$ latent tokens via cross-attention between learnable queries and the input features, and projects them to the LLM hidden dimension:
{\small
\begin{equation}
    \mathbf{H}^{m} = \text{PerceiverResampler}(\mathbf{X}^{m}) \in \mathbb{R}^{N_m \times d_h}, \quad m \in \{a, v\},
\end{equation}
}
where $d_h$ denotes the LLM hidden size.

We then construct a unified multimodal input sequence by concatenating the compressed modality tokens with special delimiter tokens and the projected textual embeddings:
\begin{equation}
    \mathbf{Z} = [\langle\text{vision}\rangle\; \mathbf{H}^{v}\;  \langle\text{audio}\rangle\; \mathbf{H}^{a}\; 
    \langle\text{text}\rangle\; \mathbf{H}^{t}\; 
    \langle\text{CLS}\rangle],
\end{equation}
where $\mathbf{H}^{t}$ denotes the projected textual hidden states, and the special tokens $\langle\text{audio}\rangle$, $\langle\text{vision}\rangle$ serve as modality boundary markers. The entire sequence is fed as input embeddings into an LLM, which performs cross-modal fusion through its self-attention layers:
\begin{equation}
    \mathbf{O} = \text{LLM}_{\phi}(\mathbf{Z}),
\end{equation}

Finally, the hidden state at the $\langle\text{CLS}\rangle$ position is extracted and passed through a classification head to obtain the predicted depression recognition label:
\begin{equation}
    \hat{y}_i^{\text{DR}} = \text{Softmax}(\text{MLP}(\mathbf{O}_{[\text{CLS}]})),
\end{equation}
where $\mathbf{O}_{[\text{CLS}]}$ denotes the LLM's last-layer hidden state at the $\langle\text{CLS}\rangle$ token position. The model is optimized with the standard cross-entropy loss:
\begin{equation}
\mathcal{L}_{\text{DR}} = - \frac{1}{N} \sum_{i=1}^{N} \sum_{k=1}^{C} y_{i,k}^{\text{DR}} \log(\hat{y}_{i,k}^{\text{DR}}),
\end{equation}
where $C$ denotes the number of recognition-label classes.

%% file: figure/framework_tex.tex
\begin{figure*}[htb]
\centering
\setlength{\belowcaptionskip}{0.2cm}
\setlength{\abovecaptionskip}{0.1cm}
\includegraphics[width=1\textwidth]{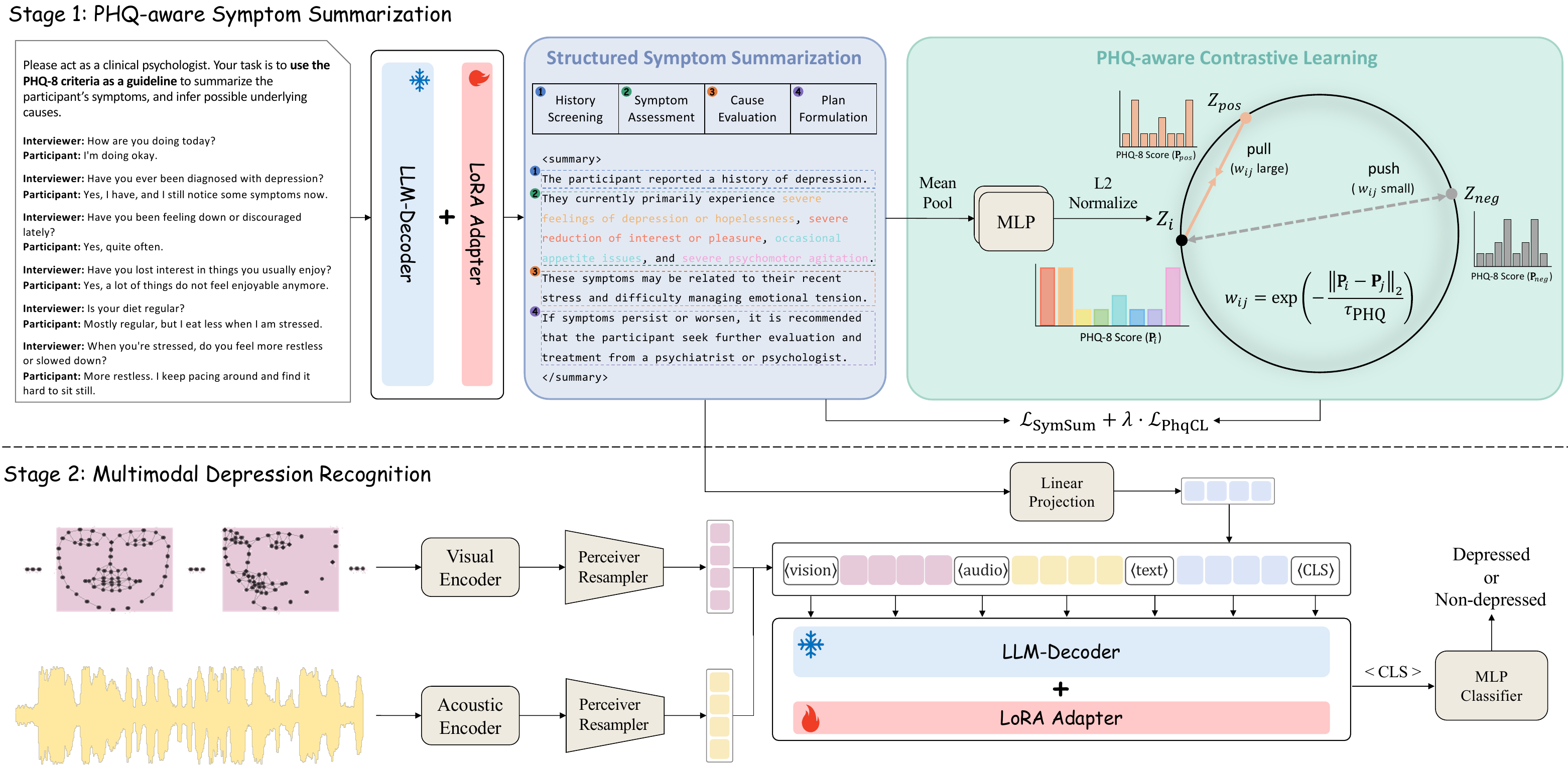}
\caption{The overview of our proposed PHQ-aware contrastive multimodal learning (PhqCML) method.}
\label{fig:framework}
\end{figure*}

%% file: section/experiments.tex
\section{Experiments}

\subsection{Evaluation Metrics}
We evaluate the framework on symptom summarization and depression recognition. \textbf{Symptom summarization.}
To evaluate the quality of generated PHQ-aware symptom summaries, we adopt standard automatic metrics widely used in text summarization~\cite{van2024adapted}. Specifically, we report ROUGE-1, ROUGE-2, and ROUGE-L to measure n-gram overlap and structural similarity, BLEU to assess surface-level precision, and BERTScore to capture semantic consistency between generated summaries and references. \textbf{Depression recognition.}
We follow previous studies~\cite{yoon2022d, zhao2025decoupled} and evaluate the model on binary classification (depressed vs. non-depressed) using precision, recall, and Macro-F1. Moreover, to ensure consistency with previous studies~\cite{niu2021hcag, chen2024depression} for experiments on the development set, we report the F1 scores for the depressed class and the control class instead of precision and recall.

\subsection{Baseline Systems}
\textbf{General-purpose LLM API baselines.}
We include five strong general-purpose LLM API baselines: \href{https://developers.openai.com/api/docs/models/gpt-5.5}{\textit{GPT-5.5}}, 
\href{https://platform.claude.com/docs/en/about-claude/models/whats-new-claude-4-8}{\textit{Claude-Opus-4.8}}, \href{https://ai.google.dev/gemini-api/docs/models/gemini-3.5-flash}{\textit{Gemini-3.5-Flash}}, \href{https://docs.z.ai/guides/llm/glm-5.2}{\textit{GLM-5.2}}, and \href{https://www.qwencloud.com/models/qwen3.7-max}{\textit{Qwen3.7-Max}}. Each API baseline is evaluated under both zero-shot and four-shot settings using the interview transcript as input. These models provide a direct reference for whether prompt-based general-purpose reasoning is sufficient for Explain-MDRC without task-specific training.

\textbf{General multimodal and MDRC baselines.} \textit{MulT}~\cite{tsai2019multimodal} is a transformer-based fusion model that uses cross-modal attention to capture interactions among unaligned modalities. \textit{MISA}~\cite{hazarika2020misa} encourages the shared space to capture cross-modal commonality while preserving private modality cues. These two methods provide strong general-purpose multimodal fusion references. \textit{D-vlog}~\cite{yoon2022d} models depressive signals from naturalistic vlog-style behavior, making it a representative depression-oriented multimodal baseline. \textit{EmoGRU}~\cite{shen2022automatic} emphasizes emotional dynamics in audio-text sequences through recurrent modeling. \textit{HiQuE}~\cite{jung2024hique} introduces hierarchical question embeddings to better exploit the structure of clinical interviews. \textit{DMPF}~\cite{zhao2025decoupled} focuses on speech-based depression detection by decoupling complementary feature perspectives before fusion. \textit{MTRB}~\cite{zhang2025optimizing} treats doctor-patient interviews as multi-instance inputs, reducing the reliance on a single global interview representation.

\textbf{LLM training and ablation baselines.}
We implement LLM-based training baselines with 
\href{https://huggingface.co/meta-llama/Llama-3.2-3B-Instruct}{\textit{Llama-3.2-3B-Instruct}} and \href{https://huggingface.co/Qwen/Qwen3-4B-Instruct-2507}{\textit{Qwen3-4B-Instruct-2507}}. \textit{DepRec} directly maps raw multimodal inputs to the depression label, testing whether an instruction-tuned LLM can perform end-to-end recognition without explicit symptom supervision. \textit{MulTsk} keeps the same two-stage symptom summarization and depression recognition pipeline as \textit{PhqCML}, but removes PHQ-aware contrastive learning; it therefore isolates the contribution of the proposed representation objective.

\textbf{Development-split references.}
Because prior DAIC-WOZ studies commonly report development-set results, we also evaluate on the development split of Explain-DAIC for direct comparison with earlier work. We report strong general-purpose LLM API references, including \href{https://qwen.ai/blog?id=fdfbaf2907a36b7659a470c77fb135e381302028&from=research.research-list}{\textit{Qwen3-Omni-30B}}, \href{https://api-docs.deepseek.com/news/news251201}{\textit{DeepSeek-v3.2}}, and \href{https://developers.openai.com/api/docs/models/gpt-5.5}{\textit{GPT-5.5}}. We also include specialized MDRC systems. \textit{$\omega$-GCN}~\cite{Burdisso2023NodeweightedGC} builds a node-weighted graph convolutional network over transcript units to exploit conversational structure. \textit{Wav2VecDep}~\cite{zhang2024improving} transfers wav2vec2.0 speech representations to depression detection. \textit{AudLadLLM}~\cite{zhang-etal-2024-llms} augments LLM reasoning with acoustic landmarks, linking speech cues with textual context. \textit{SEGA}~\cite{chen2024depression} constructs structural-element graphs for LLM-assisted depression analysis. \textit{AgentMental}~\cite{hu2026agentmental} formulates assessment as an interactive multi-agent process. \textit{PsyGAT}~\cite{vyalla2026psychologically} uses psychologically grounded graph attention to model clinically relevant cues.

\subsection{Experimental Settings}
\textbf{Symptom summarization setup.}
We select Llama-3.2-3B-Instruct and Qwen3-4B-Instruct-2507 as the LLM backbones, with the maximum input length set to 8,192 tokens and the maximum generation length set to 4,096 tokens. We employ LoRA with rank $r=16$ and scaling factor $\alpha=32$. In PhqCL, the projection head $g_{\phi}$ is a two-layer MLP with GELU activation and dropout $0.1$, mapping the LLM hidden dimension to a 128-dimensional contrastive embedding. We use a FIFO memory bank of size $M=256$ to store recent $(\mathbf{z}_j, \mathbf{p}_j)$ pairs, which is updated without gradients. The batch size is 2, the learning rate is 5e-5, and the number of training epochs is 7.
\textbf{Depression recognition setup.}
We reuse the same LLM backbones with LoRA. For the Perceiver Resampler, both audio and vision use 32 latent queries. The maximum sequence length for audio and vision inputs is set to the mean plus three standard deviations of the training set. The batch size is 4, the learning rate is 5e-5, and the number of training epochs is 5. Both stages are optimized with AdamW.
\textbf{Reproducibility.}
All experiments are conducted using PyTorch 2.6.0 with CUDA 11.8 on an NVIDIA A800 GPU. We run each trainable experiment with three random seeds and report the mean performance. The code, configurations, prompts, and processed annotation files will be released upon publication, subject to DAIC-WOZ data-use and redistribution constraints.

\input{table/main_results}

\subsection{Main Results}
Table~\ref{tab:main_results} reports the performance of different methods on the proposed Explain-DAIC dataset. Under our evaluation protocol, PhqCML achieves the strongest results across the reported metrics. The results support three main observations.

\textbf{General-purpose LLM baselines remain far behind task-specific PHQ-aware learning.}
The strongest API baseline reaches 72.99\% Macro-F1 with \textit{GPT-5.5} under four-shot prompting, which remains far below \textit{PhqCML}. This result suggests that prompting general-purpose large models alone is insufficient for reliable Explain-MDRC.

\textbf{PhqCML improves recognition while exposing intermediate evidence.}
Beyond the API comparison, \textit{PhqCML-Qwen3} reaches 95.10\% Macro-F1, outperforming the strongest prior MDRC baseline under the compared protocol (\textit{MTRB}) by 7.10 percentage points while producing PHQ-8-aligned symptom summaries alongside its predictions. Compared with single-stage LLM baselines that directly map raw dialogue to labels, \textit{PhqCML} also improves Macro-F1 by 8.71 points over \textit{DepRec-Llama3.2} and 10.23 points over \textit{DepRec-Qwen3}. These results suggest that explicit symptom summarization provides useful intermediate supervision rather than merely serving as an explanation layer.

\textbf{PHQ-aware contrastive learning consistently benefits both stages.}
Compared with \textit{MulTsk}, which uses the same two-stage pipeline without the PHQ-aware contrastive learning loss, \textit{PhqCML} improves both symptom summarization and depression recognition across Llama3.2 and Qwen3 backbones. On summarization, the largest gains appear on ROUGE-2 (+1.69 for Llama3.2 and +3.05 for Qwen3), indicating better coverage of clinically meaningful symptom phrases. The same benefit carries into recognition, where \textit{PhqCML} improves Macro-F1 by 4.45 and 2.32 points over the corresponding \textit{MulTsk} variants. Because the pipeline is otherwise unchanged, these gains isolate the effect of PHQ-aware contrastive learning: aligning summary representations with each participant's PHQ-8 profile makes the intermediate summaries more useful for downstream multimodal fusion.

\subsection{In-depth Analysis of PHQ-aware Contrastive Learning (PhqCL)}
To understand what \textit{PhqCL} contributes to \textit{PhqCML} beyond the aggregate gains reported above, we conduct representation probing, attention-shift analysis, comparison with supervised contrastive learning, and hyperparameter sensitivity analysis.

\begin{figure*}[!t]
\centering
\setlength{\belowcaptionskip}{-0.1cm}
\setlength{\abovecaptionskip}{0.2cm}
\includegraphics[width=0.9\textwidth]{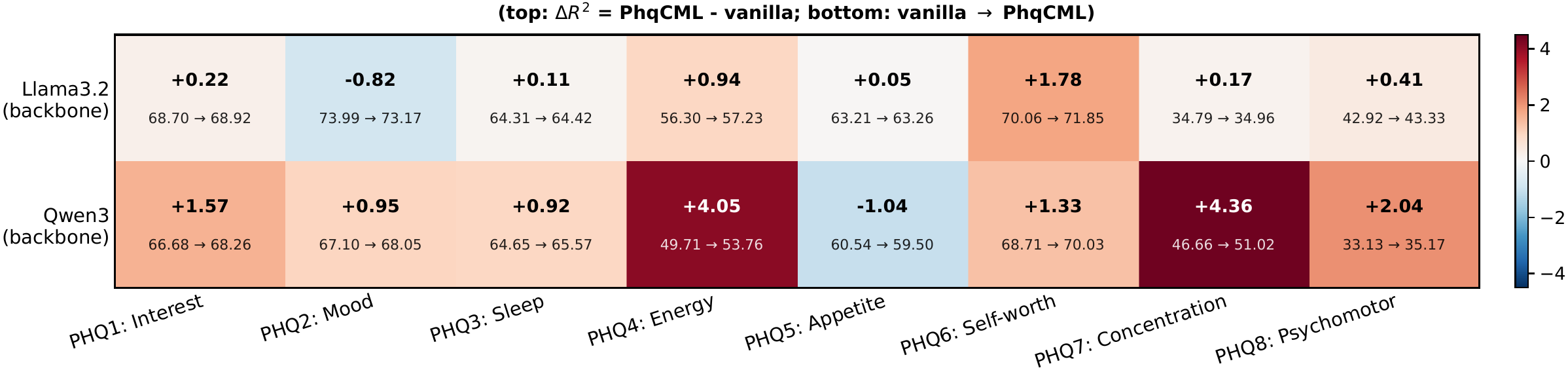}
\caption{Per-item PHQ-8 linear probing of symptom-summary embeddings from \textit{PhqCML} and its vanilla counterpart, \textit{MulTsk}. For each backbone, we train eight ridge regressors on frozen mean-pooled summary embeddings to predict individual PHQ-8 item scores on the test set. Each cell shows the change in test $R^{2}$, $\Delta R^{2}=R^{2}_{\textit{PhqCML}}-R^{2}_{\textit{MulTsk}}$, on the top, and the absolute \textit{MulTsk}$\to$\textit{PhqCML} $R^{2}$ values on the bottom. Positive values and warmer colors indicate that PHQ-aware contrastive learning makes the representations more linearly predictive of the corresponding symptom item.}
\label{fig:in-depth-PhqCL-1}
\end{figure*}

\begin{figure*}[!t]
\centering
\setlength{\belowcaptionskip}{-0.1cm}
\setlength{\abovecaptionskip}{0.2cm}
\includegraphics[width=0.93\textwidth]{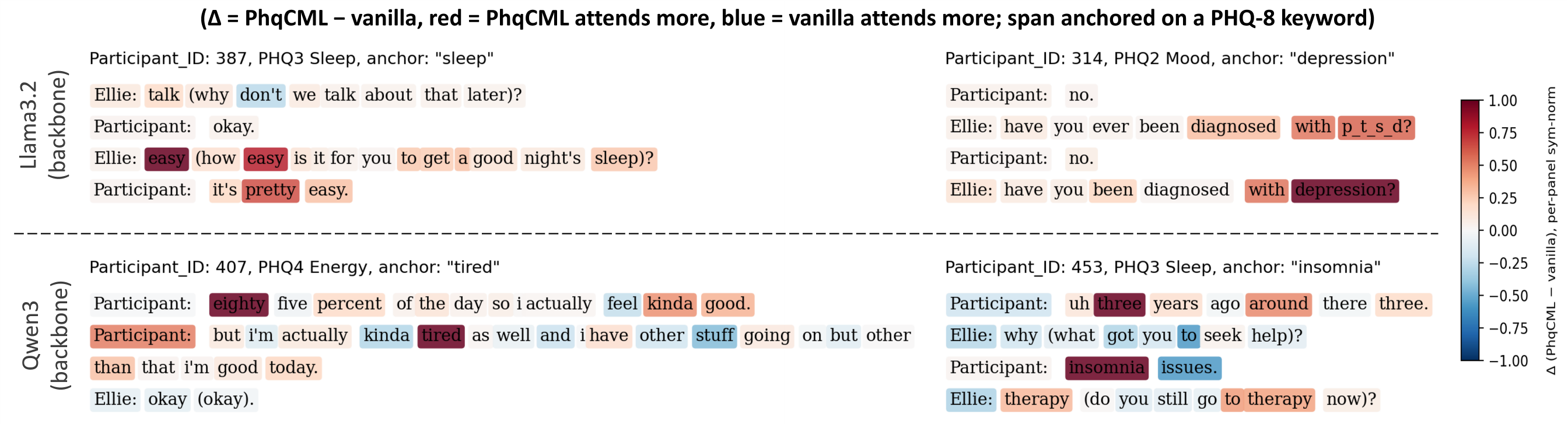}
\caption{Token-level attention shift of \textit{PhqCML} versus vanilla (\textit{MulTsk}) around PHQ-8 symptom keywords, averaged over the last four decoder layers, all heads, and all summary query positions.}
\label{fig:in-depth-PhqCL-attention}
\end{figure*}

\textbf{PhqCML improves per-item PHQ-8 probing R\textsuperscript{2} over vanilla.}
We first test whether PhqCL makes summary embeddings encode item-level symptom information. We fit eight ridge regressors from frozen summary embeddings to PHQ-8 item scores.
As shown in Fig.~\ref{fig:in-depth-PhqCL-1}, \textit{PhqCL} improves probing R\textsuperscript{2} on seven of eight items for both backbones, with mean gains of $+0.36$ and $+1.77$ percentage points for Llama3.2 and Qwen3.
The largest improvements appear on clinically salient items such as Concentration and Energy for Qwen3, suggesting that PHQ-aware supervision makes summary representations more discriminative at the symptom level.

\textbf{PhqCL redirects attention towards PHQ-8 symptom keywords.}
We then inspect whether the representation objective changes which transcript tokens the model attends to during summarization. We compare token-level attention patterns between \textit{PhqCML} and the vanilla model.
Fig.~\ref{fig:in-depth-PhqCL-attention} shows that positive attention shifts concentrate around PHQ-8 symptom anchors and nearby clinical context, while back-channel and filler tokens receive lower relative attention.
This supports the view that \textit{PhqCL} improves summary quality by reweighting transcript evidence toward symptom-relevant cues.

\begin{figure*}[htb]
\centering
\setlength{\belowcaptionskip}{-0.1cm}
\setlength{\abovecaptionskip}{0.2cm}
\includegraphics[width=0.9\textwidth]{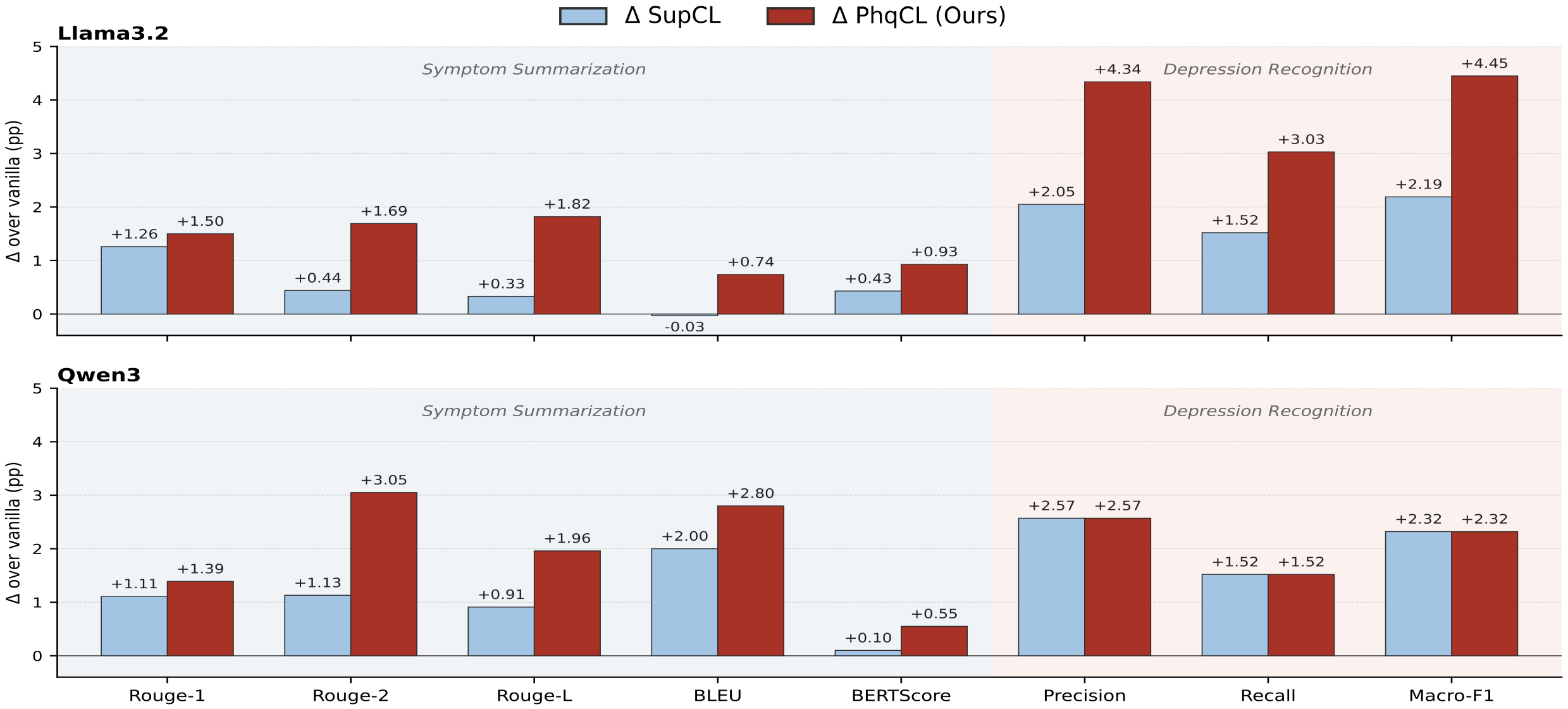}
\caption{Comparison of \textit{PhqCL} (ours) against supervised contrastive learning (\textit{SupCL}). Bars report the absolute gain ($\Delta$, in
  percentage points) over the vanilla.}
\label{fig:in-depth-PhqCL-3}
\end{figure*}

\textbf{Comparison with supervised contrastive learning (SupCL).}
To test whether continuous PHQ-8 supervision is more informative than binary class supervision, Fig.~\ref{fig:in-depth-PhqCL-3} compares \textit{PhqCL} with \textit{SupCL}, which uses binary depression labels to form hard contrastive pairs.
\textit{PhqCL} matches or exceeds \textit{SupCL} on all metrics, with clearer advantages on summary fidelity (e.g., Qwen3 ROUGE-2 $+3.05$ vs.\ $+1.13$) and Llama3.2 recognition (Macro-F1 $+4.45$ vs.\ $+2.19$).
This indicates that continuous PHQ-8 profiles provide finer clinical supervision than coarse class labels.

\begin{figure*}[htb]
\centering
\setlength{\belowcaptionskip}{-0.1cm}
\setlength{\abovecaptionskip}{0.2cm}
\includegraphics[width=0.9\textwidth]{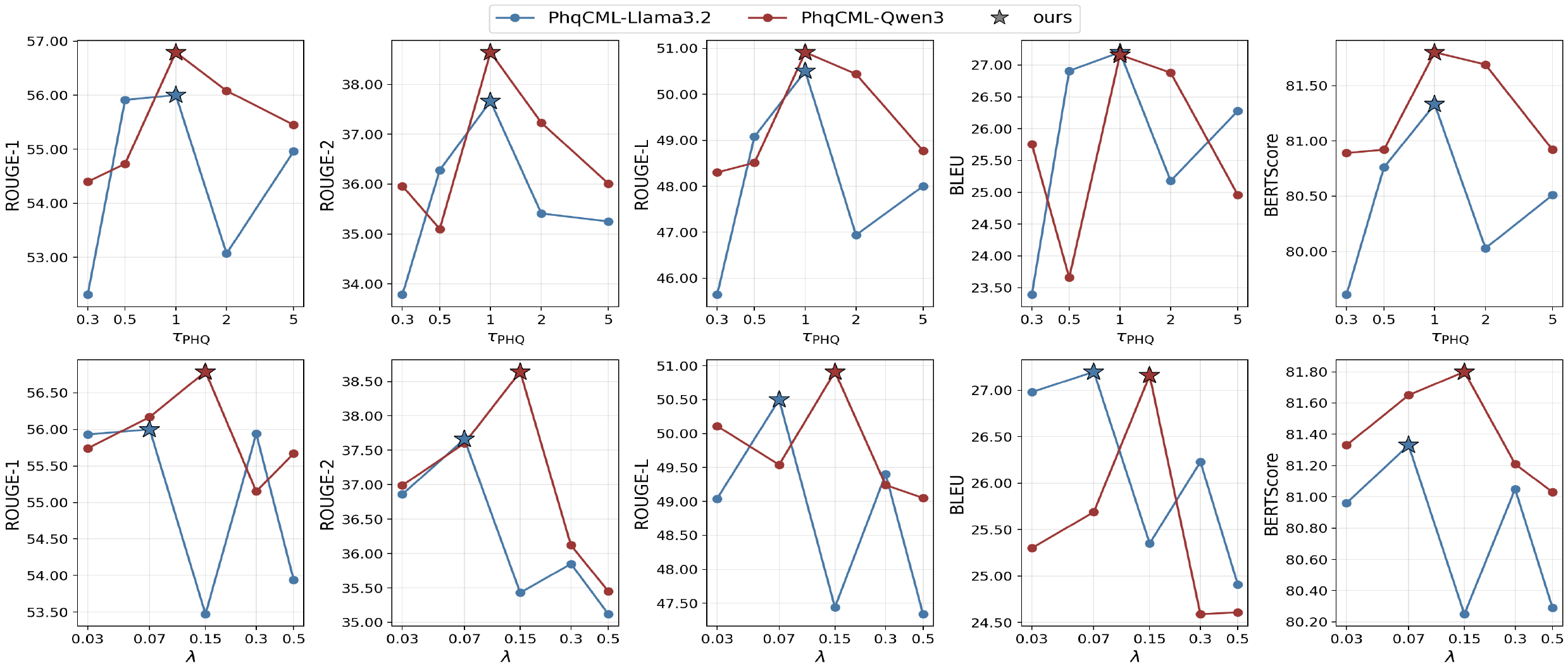}
\caption{Hyperparameter analysis of \textit{PhqCML}. Stars ($\star$) mark each backbone's chosen configuration, lying at the peak of its corresponding curve.}
\label{fig:in-depth-PhqCL-4}
\end{figure*}

\textbf{Hyperparameter analysis.}
\label{sec:ablation_phq}
Finally, we evaluate the sensitivity of PhqCL to its two key hyperparameters. Fig.~\ref{fig:in-depth-PhqCL-4} sweeps the loss weight $\lambda$ and PHQ-distance temperature $\tau_{\text{PHQ}}$.
Both backbones prefer $\tau_{\text{PHQ}}=1$, with different optimal loss weights ($0.07$ for Llama3.2 and $0.15$ for Qwen3).
Performance is stable around the selected settings, whereas overly large $\lambda$ over-regularizes generation and extreme temperatures weaken PHQ-aware similarity modeling.

\subsection{Modality Ablation Study}

\input{table/modality_ablation}
To evaluate the contribution of each input modality, we conduct ablation studies on both \textit{PhqCML-Llama3.2} and \textit{PhqCML-Qwen3} by progressively removing acoustic and visual modalities (Table~\ref{tab:modality_ablation}). Removing audio or vision each causes a Macro-F1 drop of 2.32 points for Qwen3 and 2.26 points for Llama3.2; removing both modalities leads to drops of 4.58 and 4.45 points, respectively.
These consistent degradations suggest that nonverbal modalities provide complementary cues (prosody, vocal quality, and facial expression) beyond the textual stream for depression recognition.

\subsection{Performance on the DAIC-WOZ Development Split}
\input{table/dev_eval}
Since Explain-DAIC is derived from DAIC-WOZ, where prior studies report on the development split, we evaluate \textit{PhqCML} on this split for direct comparison (Table~\ref{tab:dev_set_comparison}).
General-purpose LLMs lag behind specialized MDRC models: even \textit{GPT-5.5} reaches only 76.31 Macro-F1, 13 points below \textit{PsyGAT} (89.99), suggesting that prompting alone is insufficient for robust depression recognition.
Both \textit{PhqCML} variants achieve the strongest reported results: \textit{PhqCML-Llama3.2} and \textit{PhqCML-Qwen3} each reach 93.88 Macro-F1, surpassing \textit{PsyGAT} by 3.89 points.
Beyond this quantitative gain, \textit{PhqCML} generates a structured PHQ-8--aligned symptom summary alongside each prediction, providing clinician-readable intermediate evidence that prior classification-only methods lack.

\input{table/cmdc_generalization}

\subsection{Cross-dataset Depression Prediction via Dialogue-based PHQ-style Assessment}

To examine whether the dialogue-based symptom-assessment ability of \textit{PhqCML} transfers beyond Explain-DAIC, we conduct an external cross-dataset experiment on CMDC~\cite{zou2022semi}, a Mandarin PHQ-9-based semi-structured clinical-interview corpus for binary depression recognition. 
Unlike conventional depression-recognition models that directly map an interview to a binary label, \textit{PhqCML-SymSum} first performs PHQ-style symptom assessment from the dialogue transcript by generating a structured symptom summary and estimating item-level symptom scores. 
The depression label can then be inferred from the estimated scale profile. This design allows us to test whether the model has learned transferable clinical-scale-guided reasoning rather than only dataset-specific label cues.

CMDC differs from Explain-DAIC in language, interview protocol, and the acquisition and feature format of acoustic and visual modalities. 
As a result, the second-stage multimodal depression-recognition head trained on Explain-DAIC cannot be directly applied to CMDC. 
We therefore focus this experiment on the first stage of \textit{PhqCML}, namely \textit{PhqCML-SymSum}, and evaluate whether its dialogue-based PHQ assessment ability can transfer to CMDC transcripts.

Specifically, we translate the Mandarin CMDC transcripts into English and apply the Explain-DAIC-trained \textit{PhqCML-SymSum} without any CMDC-specific fine-tuning. 
Given only the interview transcript, the model generates a structured symptom summary and estimates PHQ-style item scores. 
Cases with an estimated total score greater than 9 are classified as depressed, following the commonly used PHQ cutoff for moderate depressive symptoms~\cite{kroenke2001phq}.
Because Explain-DAIC is annotated with PHQ-8 whereas CMDC follows PHQ-9, we further construct \textit{PhqCML-SymSum}$^{*}$ by adding a prompt-level instruction for the PHQ-9 suicidality item. This variant adapts the output format to the PHQ-9 setting but does not retrain the model on CMDC.

Table~\ref{tab:cmdc_generalization} compares our method with recent strong CMDC baselines. 
Without CMDC-specific fine-tuning, \textit{PhqCML-SymSum-Qwen3} achieves 92.86 Macro-F1, which is comparable to \textit{IIFDD} with 93.00 Macro-F1. 
After incorporating the suicidality item, \textit{PhqCML-SymSum-Qwen3}$^{*}$ reaches 95.79 Macro-F1 and 97.12 Recall, narrowing the gap to CMDC-specialized multimodal systems such as \textit{VT-MSC}. These results suggest that \textit{PhqCML-SymSum} can transfer its dialogue-based PHQ-style assessment ability across datasets, languages, and screening scales. Rather than directly fitting a CMDC-specific classifier, the model infers depression labels through an intermediate symptom-score profile, providing a more clinically structured prediction pathway. 

\subsection{Clinical Expert Evaluation}
Automatic metrics may not fully capture whether generated summaries are clinically readable and evidence-grounded. We therefore recruited three licensed psychiatrists for a blinded paired preference evaluation on the full Explain-DAIC test set. Following prior work~\cite{van2024adapted}, the experts compared anonymized summaries generated by \textit{PhqCML-Qwen3} and \textit{GPT-5.5} along three clinically oriented dimensions: completeness, correctness, and conciseness. Each paired comparison was accompanied by the corresponding interview transcript, allowing experts to assess whether the generated symptom descriptions were supported by the source dialogue. \textbf{Completeness} measures whether the summary captures the major symptom domains and contextual evidence expressed in the interview, rather than only repeating the most explicit statements.
\textbf{Correctness} focuses on factual and clinical faithfulness, including whether symptom descriptions are grounded in the participant's dialogue and avoid over-interpreting ambiguous cues.
\textbf{Conciseness} evaluates whether the summary suppresses redundant, off-topic, or weakly supported details while preserving information needed for assessment.

For each test case, the two system outputs were presented as paired summaries without revealing model identities, and each psychiatrist independently assigned a preference score from $-2$ to $2$. The psychiatrists were instructed to base their judgments only on the source dialogue and the anonymized summaries, focusing on whether the generated symptom descriptions were supported by the interview evidence and clinically useful for review.
Positive scores indicate that \textit{PhqCML-Qwen3} is preferred, negative scores indicate that \textit{GPT-5.5} is preferred, and zero denotes no clear preference.
For statistical testing, we first averaged the three psychiatrists' preference scores for each test case and each evaluation dimension, and then applied a one-sample Wilcoxon signed-rank test to the resulting case-level mean preference scores against zero. Results are reported as mean $\pm$ SEM.
This design complements automatic overlap metrics by asking whether the generated explanations satisfy practical clinical requirements: a useful summary should cover the relevant PHQ-8 symptom evidence, avoid unsupported symptom attributions, and remain concise enough for efficient review.

\input{figure/clinical_evaluation_violin}
\input{table/clinicial_evaluation}

Under this clinically oriented protocol, the results in Table~\ref{tab:clinical_evaluation} show consistently positive expert preferences for \textit{PhqCML} over \textit{GPT-5.5} in completeness, correctness, and conciseness ($p<0.001$).
The largest margin appears in correctness, the dimension that most directly tests whether symptom descriptions are grounded in interview evidence and avoid unsupported clinical inference.
This indicates that the expert preference is not merely driven by longer or more fluent summaries, but by explanations that more faithfully connect patient-reported evidence with PHQ-8 symptom interpretations.
These findings support the central motivation of Explain-MDRC: explicit PHQ-aware training improves the clinical readability and evidence grounding of the intermediate explanation, while also improving downstream recognition scores.

Fig.~\ref{fig:clinical_evaluation_violin} further examines the distribution of expert preferences for each psychiatrist, complementing the aggregate case-level test with a per-expert view.
Across the three psychiatrists, positive preferences for \textit{PhqCML} account for 88.0\% of judgments on completeness, 93.6\% on correctness, and 74.5\% on conciseness.
The distributions are concentrated in the ``Good'' and ``Very Good'' categories, especially for correctness, where very few cases favor \textit{GPT-5.5}.
This pattern suggests that \textit{PhqCML} more consistently avoids hallucinated or clinically unsupported symptom descriptions, which is important for clinician-reviewable mental-health assessment tools.
Conciseness shows a slightly broader distribution than the other two dimensions, indicating that some generated summaries may still include nonessential details. Nevertheless, the overall preference remains positive, suggesting that the structured PHQ-aware formulation improves focus without sacrificing symptom coverage.

%% file: table/main_results.tex
\begin{table*}[!t]
\centering
\small
\setlength{\belowcaptionskip}{0.1cm}
\setlength{\abovecaptionskip}{0.2cm}
\caption{\label{tab:main_results}
Comparison results on the Explain-DAIC dataset. ``--'' indicates that the corresponding method does not produce symptom summaries. The best results are marked in bold.
}
\scalebox{0.88}{
\setlength{\tabcolsep}{3.4mm}
\begin{tabular}{llcccccccc}
\toprule
\multicolumn{2}{l}{\multirow{2}{*}{\textbf{Methods}}} & \multicolumn{5}{c}{Symptom Summarization}  & \multicolumn{3}{c}{Depression Recognition}  \\
\multicolumn{2}{l}{}   & ROUGE-1    & ROUGE-2     & ROUGE-L   & BLEU            & BERTScore      & Precision  & Recall   & Macro-F1    \\ \midrule
\multicolumn{2}{l}{GPT-5.5 (zero-shot)} & 38.34 & 18.43 & 28.50 & 15.85 & 73.89 & 52.38 & 78.57 & 70.41 \\
\multicolumn{2}{l}{Claude-Opus-4.8 (zero-shot)} &29.63  &12.79  &20.73 &8.67  &68.92  &66.67  &42.86 &68.34  \\
\multicolumn{2}{l}{Gemini-3.5-Flash (zero-shot)} &43.65  &24.48  &35.86  &19.45  & 76.92 &60.00  &42.86  &66.43  \\
\multicolumn{2}{l}{GLM-5.2 (zero-shot)} &33.51  &15.44  &23.99  &11.54  &70.65  &66.67  &42.86  &68.34  \\
\multicolumn{2}{l}{Qwen3.7-Max (zero-shot)} &43.20  &23.11  &34.75  &21.52  &75.83  &70.00  &50.00  &72.02  \\
\multicolumn{2}{l}{GPT-5.5 (four-shot)} &42.68  &21.36  &33.07  &19.66  &76.28  &54.55  &85.71  &72.99  \\
\multicolumn{2}{l}{Claude-Opus-4.8 (four-shot)} &43.02  &22.93  &33.13  &18.24  &75.44  &66.67  &42.86  &68.34  \\
\multicolumn{2}{l}{Gemini-3.5-Flash (four-shot)} &49.21  &30.47  &42.03  &22.85  &78.90  &66.67  &42.86  &68.34  \\
\multicolumn{2}{l}{GLM-5.2 (four-shot)} &43.69  &24.52  &34.44  &20.76  &75.66  &62.50  &35.71  &64.39  \\
\multicolumn{2}{l}{Qwen3.7-Max (four-shot)} &49.84  &30.20  &42.10  &23.78  &78.39  &63.64  &50.00  &70.03  \\
\multicolumn{2}{l}{MulT (ACL' 2019)~\cite{tsai2019multimodal}}  &  --   &   --   &   --   &  --  &  --  & 73.00       & 74.00  & 74.00  \\
\multicolumn{2}{l}{MISA (ACM MM' 2020)~\cite{hazarika2020misa}}  &  --   &   --   &   --   &  --  &  --  & 74.00       & 77.00  & 74.00  \\
\multicolumn{2}{l}{D-vlog (AAAI' 2020)~\cite{yoon2022d}}  &  --   &   --   &   --   &  --  &  --  & 73.00       & 72.00  & 73.00  \\
\multicolumn{2}{l}{EmoGRU (ICASSP' 2022)~\cite{shen2022automatic}}  &  --   &   --   &   --   &  --  &  --  & 75.00      & 78.00    & 75.00 \\
\multicolumn{2}{l}{HiQuE (CIKM' 2024)~\cite{jung2024hique}}  &  --   &   --   &   --   &  --  &  --  & 78.00     & 80.00  & 79.00 \\
\multicolumn{2}{l}{DMPF (TAFFC' 2025)~\cite{zhao2025decoupled}}  &  --   &   --   &   --   &  --  &  --  & 79.92     & 81.30  & 80.30 \\
 \multicolumn{2}{l}{MTRB (Sci. Rep. 2025)~\cite{zhang2025optimizing}}  &  --   &   --   &   --   &  --  &  --  & 89.00     & 86.00  & 88.00 \\
\multicolumn{2}{l}{DepRec-Llama3.2}  &  --   &   --   &   --   &  --  &  --  & 83.33  & 89.39  & 84.07 \\
\multicolumn{2}{l}{DepRec-Qwen3}  &  --   &   --   &   --   &  --  &  --  &  83.93   & 86.77 & 84.87 \\
\multicolumn{2}{l}{MulTsk-Llama3.2}     &  54.50      &  35.97   & 48.73    &  26.46   & 80.40  & 86.84    &  92.42  & 88.33 \\
\multicolumn{2}{l}{MulTsk-Qwen3}   &  55.40   & 35.59      & 48.95   & 24.36       &  81.25  & 91.18  & 95.45  & 92.78  \\
\multicolumn{2}{l}{\textbf{PhqCML-Llama3.2 (Ours)}} & 56.00 & 37.66 & 50.50  &\textbf{27.20} & 81.33 & 91.18  & 95.45  & 92.78 \\
\multicolumn{2}{l}{\textbf{PhqCML-Qwen3 (Ours)}} & \textbf{56.79} &\textbf{38.64} &\textbf{50.91} &27.16 & \textbf{81.80}  & \textbf{93.75} & \textbf{96.97} & \textbf{95.10} \\
\bottomrule
\end{tabular}
}
\end{table*}

%% file: table/modality_ablation.tex
\begin{table}[htb]
\centering
\setlength{\belowcaptionskip}{-0.2cm}
\setlength{\abovecaptionskip}{0.1cm}
\caption{\label{tab:modality_ablation}%
Modality ablation of PhqCML.
}
\setlength{\tabcolsep}{6pt}
\small
\begin{tabular}{lccc}
\toprule
\multirow{2}{*}{Methods} & \multicolumn{3}{c}{Depression Recognition} \\
 & Precision & Recall & Macro-F1 \\ \midrule
\textbf{PhqCML-Llama3.2}    & \textbf{91.18} & \textbf{95.45} & \textbf{92.78} \\
\quad -- w/o Audio          & 88.89          & 93.94          & 90.52          \\
\quad -- w/o Vision         & 88.89          & 93.94          & 90.52          \\
\quad -- w/o Audio, Vision  & 86.84          & 92.42          & 88.33          \\
\midrule
\textbf{PhqCML-Qwen3}       & \textbf{93.75} & \textbf{96.97} & \textbf{95.10} \\
\quad -- w/o Audio          & 91.18          & 95.45          & 92.78          \\
\quad -- w/o Vision         & 91.18          & 95.45          & 92.78          \\
\quad -- w/o Audio, Vision  & 88.89          & 93.94          & 90.52          \\
\bottomrule
\end{tabular}
\vspace{0pt}
\end{table}

%% file: table/dev_eval.tex
\begin{table}[!htbp]
\centering
\setlength{\belowcaptionskip}{-0.2cm}
\setlength{\abovecaptionskip}{0.1cm}
\caption{\label{tab:dev_set_comparison}%
Comparison against baselines on the Explain-DAIC development set. ``Depre'' / ``Contr'' denote F1 scores for the positive (depressed) and negative (control) classes.
}
\setlength{\tabcolsep}{5pt}
\small
\begin{tabular}{lccc}
\toprule
\multirow{2}{*}{Methods} & \multicolumn{3}{c}{Depression Recognition} \\
 & Depre & Contr & Macro-F1 \\ \midrule
 Qwen3-Omni-30B (four-shot) & 65.00 & 74.07 & 69.54 \\
DeepSeek-v3.2 (four-shot)  & 68.75 & 70.59 & 69.67 \\
GPT-5.5 (four-shot)          & 68.75 & 83.75 & 76.31 \\
$\omega$-GCN (INTERSPEECH'23)~\cite{Burdisso2023NodeweightedGC} & 78.26 & 89.36 & 83.81 \\
Wav2VecDep (Sci.Rep.'24)~\cite{zhang2024improving}    & --    & --    & 79.00 \\
AudLadLLM (EMNLP'24)~\cite{zhang-etal-2024-llms}   & --    & --    & 83.30 \\
SEGA (NAACL'24)~\cite{chen2024depression}          & 84.62 & 90.91 & 87.76 \\
AgentMental (AAAI'26)~\cite{hu2026agentmental}    & 80.00 & 92.00 & 86.00 \\
PsyGAT (arXiv'26)~\cite{vyalla2026psychologically} & 84.96 & 93.02 & 89.99 \\
\textbf{PhqCML-Llama3.2 (Ours)} & \textbf{92.31} & \textbf{95.45} & \textbf{93.88} \\
\textbf{PhqCML-Qwen3 (Ours)}    & \textbf{92.31} & \textbf{95.45} & \textbf{93.88} \\
\bottomrule
\end{tabular}
\end{table}

%% file: table/cmdc_generalization.tex
\begin{table}[!t]
\centering
\setlength{\belowcaptionskip}{0.0cm}
\setlength{\abovecaptionskip}{0.1cm}
\caption{\label{tab:cmdc_generalization}%
Cross-dataset depression prediction on CMDC by inferring PHQ-style symptom scores from dialogue.
}
\setlength{\tabcolsep}{3pt}
\small
\begin{tabular}{lccc}
\toprule
\multirow{2}{*}{Methods} & \multicolumn{3}{c}{Depression Recognition} \\
 & Precision & Recall & Macro-F1 \\ \midrule
IIFDD (Inf.Fus.'24)~\cite{chen2024iifdd} &95.00 &93.00 &93.00 \\
VT-MSC (Sci.Rep.'25)~\cite{xu2025depression} & \textbf{97.52} & 96.64 & 97.08 \\
DepressInstruct (Inf.Fus.'26)~\cite{li2026depressinstruct} & -- & -- & \textbf{98.18} \\

\textbf{PhqCML-SymSum-Llama3.2} & 90.23 & 89.42 & 89.80\\
\textbf{PhqCML-SymSum-Llama3.2$^{*}$} &91.35 &91.35 &91.35 \\
\textbf{PhqCML-SymSum-Qwen3} &92.48  &93.27 & 92.86 \\
\textbf{PhqCML-SymSum-Qwen3$^{*}$} & 94.83 & \textbf{97.12} & 95.79 \\
\bottomrule
\end{tabular}
\end{table}

%% file: figure/clinical_evaluation_violin.tex
\begin{figure*}[htb]
\centering
\setlength{\belowcaptionskip}{0.2cm}
\setlength{\abovecaptionskip}{0.1cm}
\scalebox{0.9}{
\begin{tikzpicture}[x=0.032cm,y=0.032cm,font=\scriptsize]
\definecolor{clinicalaxis}{RGB}{64,70,78}
\definecolor{expertone}{RGB}{70,120,176}
\definecolor{experttwo}{RGB}{174,52,43}
\definecolor{expertthree}{RGB}{145,145,145}

\newcommand{\stackbar}[4]{
    \pgfmathsetmacro{\barleft}{#1-10}
    \pgfmathsetmacro{\barright}{#1+10}
    \pgfmathsetmacro{\midone}{#2}
    \pgfmathsetmacro{\midtwo}{#2+#3}
    \pgfmathsetmacro{\total}{#2+#3+#4}
    \ifdim #2pt>0pt
        \draw[fill=expertone, draw=white, line width=0.35pt] (\barleft,0) rectangle (\barright,\midone);
    \fi
    \ifdim #3pt>0pt
        \draw[fill=experttwo, draw=white, line width=0.35pt] (\barleft,\midone) rectangle (\barright,\midtwo);
    \fi
    \ifdim #4pt>0pt
        \draw[fill=expertthree, draw=white, line width=0.35pt] (\barleft,\midtwo) rectangle (\barright,\total);
    \fi
}

\newcommand{\drawclinicalpanel}[3]{
    \begin{scope}[shift={({#1},0)}]
    \node[font=\small, text=black] at (75,116) {#2};
    \draw[draw=clinicalaxis, line width=0.45pt] (0,0) -- (158,0);
    \draw[draw=clinicalaxis, line width=0.45pt] (0,0) -- (0,104);
    \foreach \y in {20,40,60,80,100} {
        \draw[gray!35, dotted, line width=0.35pt] (0,\y) -- (158,\y);
        \node[anchor=east, text=clinicalaxis] at (-4,\y) {\y};
    }
    \node[anchor=east, text=clinicalaxis] at (-4,0) {0};
    \node[rotate=90, font=\small, text=black] at (-22,52) {Frequency};
    \node[font=\small, text=black] at (79,-38) {Expert preference};
    \foreach \x/\lab in {16/Very Poor,47/Poor,78/Neutral,109/Good,140/Very Good} {
        \node[anchor=east, rotate=24, text=black] at (\x+6,-5) {\lab};
    }
    #3
    \end{scope}
}

\node[font=\normalsize, text=black] at (264,136) {\textit{PhqCML-Qwen3} vs. \textit{GPT-5.5}};

\drawclinicalpanel{0}{Completeness}{
    \stackbar{16}{0}{1}{0}
    \stackbar{47}{2}{3}{1}
    \stackbar{78}{4}{4}{2}
    \stackbar{109}{17}{17}{14}
    \stackbar{140}{24}{22}{30}
}

\drawclinicalpanel{185}{Correctness}{
    \stackbar{16}{0}{0}{0}
    \stackbar{47}{1}{1}{0}
    \stackbar{78}{3}{2}{2}
    \stackbar{109}{16}{14}{13}
    \stackbar{140}{27}{30}{32}
}

\drawclinicalpanel{370}{Conciseness}{
    \stackbar{16}{0}{0}{0}
    \stackbar{47}{7}{7}{1}
    \stackbar{78}{10}{11}{0}
    \stackbar{109}{14}{14}{2}
    \stackbar{140}{16}{15}{44}
}

\begin{scope}[shift={(155,-54)}]
    \draw[fill=expertone, draw=none] (0,0) rectangle (10,6);
    \node[anchor=west, text=clinicalaxis] at (13,3) {Expert 1};
    \draw[fill=experttwo, draw=none] (72,0) rectangle (82,6);
    \node[anchor=west, text=clinicalaxis] at (85,3) {Expert 2};
    \draw[fill=expertthree, draw=none] (144,0) rectangle (154,6);
    \node[anchor=west, text=clinicalaxis] at (157,3) {Expert 3};
\end{scope}
\end{tikzpicture}
}
\caption{Per-expert frequency distributions of preference ratings for \textit{PhqCML-Qwen3} against \textit{GPT-5.5} across completeness, correctness, and conciseness. Stacked bars show the frequency of ratings assigned by each expert. Preference scores range from $-2$ (Very Poor; strongly favoring \textit{GPT-5.5}) to $2$ (Very Good; strongly favoring \textit{PhqCML-Qwen3}).}
\label{fig:clinical_evaluation_violin}
\end{figure*}

%% file: table/clinicial_evaluation.tex
\begin{table}[!t]
\centering
\small
\setlength{\belowcaptionskip}{0.1cm}
\setlength{\abovecaptionskip}{0.1cm}
\setlength{\tabcolsep}{1.6mm}
\caption{Expert preference scores for PhqCML-Qwen3 against GPT-5.5 (mean $\pm$ SEM). $^{**}$ denotes $p<0.001$ using a one-sample Wilcoxon signed-rank test on case-level mean preference scores.}
\label{tab:clinical_evaluation}
\scalebox{0.845}{
\begin{tabular}{lccc}
\toprule
\textbf{Models} & \textbf{Completeness} & \textbf{Correctness} & \textbf{Conciseness} \\
\midrule
\textit{PhqCML} vs. \textit{GPT-5.5} & \colorbox[HTML]{B0D4C1}{$1.35 \pm 0.14$}$^{**}$ & \colorbox[HTML]{B0D4C1}{$1.54 \pm 0.08$}$^{**}$ & \colorbox[HTML]{B0D4C1}{$1.17 \pm 0.12$}$^{**}$ \\
\bottomrule
\end{tabular}
}
\end{table}

%% file: section/discussion.tex
\section{Limitations}
Despite these promising results, several limitations remain. \textbf{First}, Explain-MDRC does not yet incorporate external clinical knowledge sources, such as psychiatric knowledge graphs or ontologies, which may improve the credibility and domain alignment of symptom interpretation. 
\textbf{Second}, the current framework follows a static learning paradigm; future work should explore continuous learning or human-in-the-loop adaptation so clinicians can verify outputs and provide corrective feedback, consistent with human-centered XAI principles that emphasize target-user understanding and calibrated trust~\cite{rong2024human}. 
\textbf{Third}, real-world deployment requires broader safety evaluation, including bias, unintended output, spurious contextual shortcuts, and misuse monitoring in high-stakes clinical contexts~\cite{yang2024context}. Accordingly, the current framework should be interpreted as a screening-oriented research prototype rather than a deployable diagnostic system; prospective clinical validation with clinician-adjudicated diagnostic labels is required before real-world use.
\textbf{Finally}, although this work anchors explainability around PHQ-8, extending the framework to multiple validated scales, other mental disorders, and multilingual or culturally diverse settings would improve generalizability and equitable access~\cite{first2004clinical}.

%% file: section/conclusion.tex
\section{Conclusion}

This work presents Explain-MDRC, an explainable framework that reorients multimodal depression recognition in clinical interviews from direct label prediction toward PHQ-8-aligned symptom reasoning.
By constructing Explain-DAIC and introducing structured symptom summaries as an intermediate representation, Explain-MDRC explicitly links interview evidence to clinically meaningful symptom dimensions before integrating acoustic and visual cues for final recognition.
PhqCML instantiates this framework with PHQ-aware contrastive learning, which shapes summary representations according to item-level clinical profiles and provides clinically structured evidence for multimodal recognition.
Experiments and clinical expert evaluations show that this design improves recognition performance while providing structured, interpretable symptom-level evidence. These results support clinically structured intermediate reasoning as a practical direction for transparent AI-assisted depression recognition.